\documentclass[sigconf,authorversion,nonacm]{acmart}
\AtBeginDocument{%
  }

\setcopyright{acmlicensed}
\copyrightyear{2018}
\acmYear{2018}
\acmDOI{XXXXXXX.XXXXXXX}
\acmConference[Conference acronym 'XX]{Make sure to enter the correct
  conference title from your rights confirmation email}{June 03--05,
  2018}{Woodstock, NY}
\acmISBN{978-1-4503-XXXX-X/2018/06}

\usepackage{multirow}

\usepackage{booktabs}
\usepackage{algpseudocode}
\usepackage{amsmath}
\usepackage{algorithm}
\usepackage{newfloat}
\usepackage{subcaption}

\newcommand{\eg}{{e.g.}}
\newcommand{\ie}{{i.e.}}

\newcommand{\vs}{{vs. }}

\DeclareFloatingEnvironment[
    fileext=los,
    listname={List of Prompts},
    name=Prompt,
    placement=tbhp,
]{prompt}

\begin{document}

\title{Knowledge Extraction on Semi-Structured Content: Does It Remain Relevant for Question Answering in the Era of LLMs?}

\author{Kai Sun}
\affiliation{
\institution{Meta Reality Labs}
\country{USA}
}
\authornote{Correspondence to: Kai Sun (sunkaicn@meta.com).}
\author{Yin Huang}
\affiliation{
\institution{Meta Reality Labs}
\country{USA}
}
\author{Srishti Mehra}
\affiliation{
\institution{Meta Reality Labs}
\country{USA}
}
\author{Mohammad Kachuee}
\affiliation{
\institution{Meta Reality Labs}
\country{USA}
}
\author{Xilun Chen}
\affiliation{
\institution{FAIR, Meta}
\country{USA}
}
\author{Renjie Tao}
\affiliation{
\institution{Meta Reality Labs}
\country{USA}
}
\author{Zhaojiang Lin}
\affiliation{
\institution{Meta Reality Labs}
\country{USA}
}
\author{Andrea Jessee}
\affiliation{
\institution{Meta Reality Labs}
\country{USA}
}
\author{Nirav Shah}
\affiliation{
\institution{Meta Reality Labs}
\country{USA}
}
\author{Alex Betty}
\affiliation{
\institution{Meta Reality Labs}
\country{USA}
}
\author{Yue Liu}
\affiliation{
\institution{Meta Reality Labs}
\country{USA}
}
\author{Anuj Kumar}
\affiliation{
\institution{Meta Reality Labs}
\country{USA}
}
\author{Wen-tau Yih}
\affiliation{
\institution{FAIR, Meta}
\country{USA}
}
\author{Xin Luna Dong}
\affiliation{
\institution{Meta Reality Labs}
\country{USA}
}

\renewcommand{\shortauthors}{Sun et al.}

\begin{abstract}

The advent of Large Language Models (LLMs) has significantly advanced web-based Question Answering (QA) systems over semi-structured content, raising questions about the continued utility of knowledge extraction for question answering. This paper investigates the value of triple extraction in this new paradigm by extending an existing benchmark with knowledge extraction annotations and evaluating commercial and open-source LLMs of varying sizes. Our results show that web-scale knowledge extraction remains a challenging task for LLMs. Despite achieving high QA accuracy, LLMs can still benefit from knowledge extraction, through augmentation with extracted triples and multi-task learning. These findings provide insights into the evolving role of knowledge triple extraction in web-based QA and highlight strategies for maximizing LLM effectiveness across different model sizes and resource settings.

\end{abstract}



\maketitle

\section{Introduction} %
\label{sec:intro}
Over the past decade, Knowledge Graphs (KGs) have played a pivotal role in advancing Question Answering (QA) systems, powering major industrial applications such as Google and Bing's search~\cite{zhu2020asurvey}. Closely tied to this progress is the field of Information Extraction, which focuses on deriving structured knowledge---typically in the form of triples---from unstructured or semi-structured text~\cite{srihari1999information,khot-etal-2017-answering}. Recently, Large Language Models (LLMs) have shown remarkable proficiency in interpreting unstructured content and generating accurate responses directly from it, a capability central to techniques like \textit{Retrieval-Augmented Generation (RAG)}~\cite{lewis2020retrieval}. In light of these developments, an important question arises: \textit{in an era dominated by powerful generative models, does the task of structured knowledge extraction still hold relevance for question answering?}

\begin{figure*}[h!]
\centering
    \begin{minipage}{0.665\textwidth}%
        \begin{subfigure}{0.36\linewidth}%
            \includegraphics[width=\textwidth]{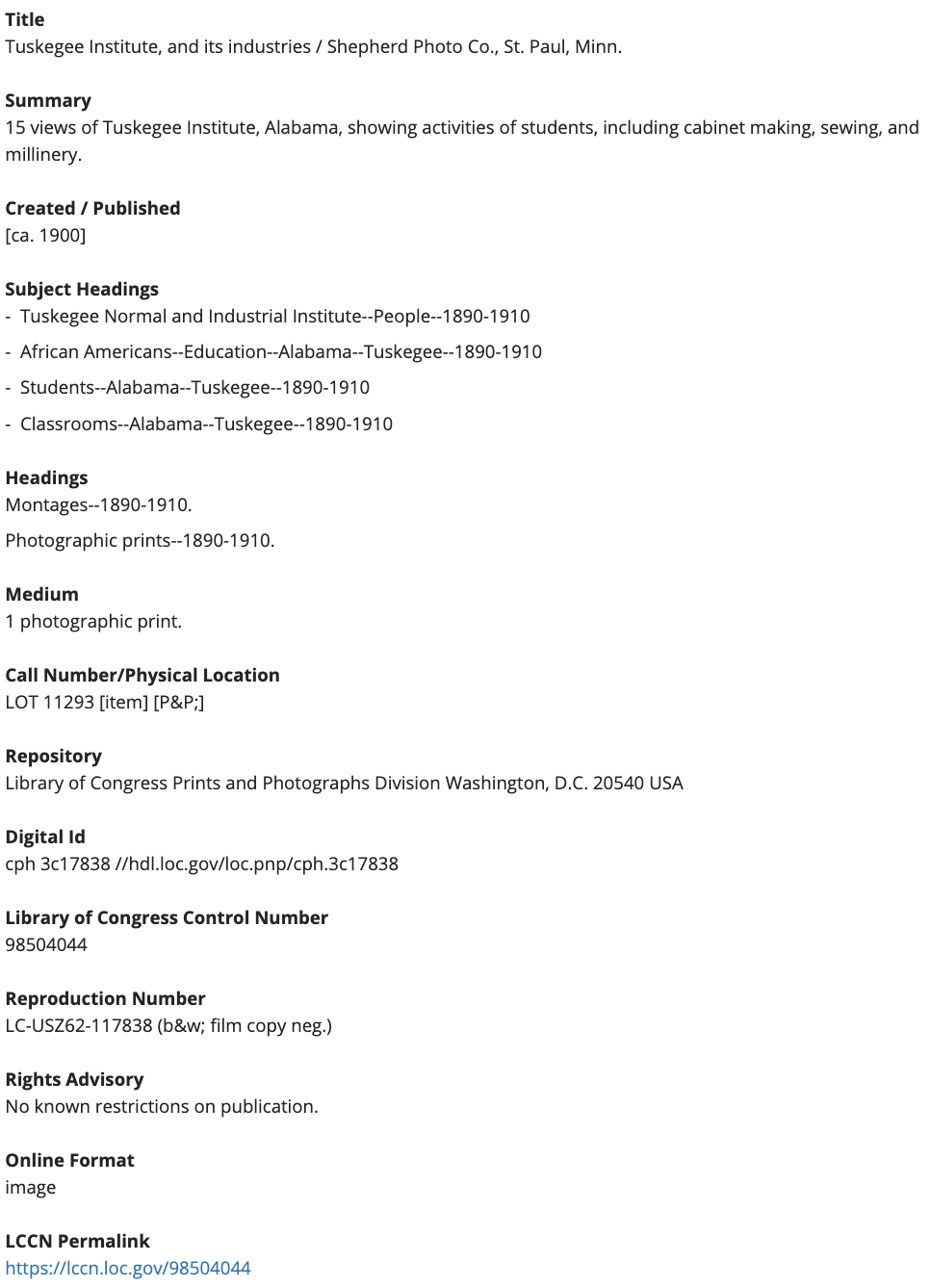} %
        \caption{}
        \label{fig:attributevalue}
        \end{subfigure}
        \begin{subfigure}{0.625\linewidth}%
            \includegraphics[width=\linewidth]{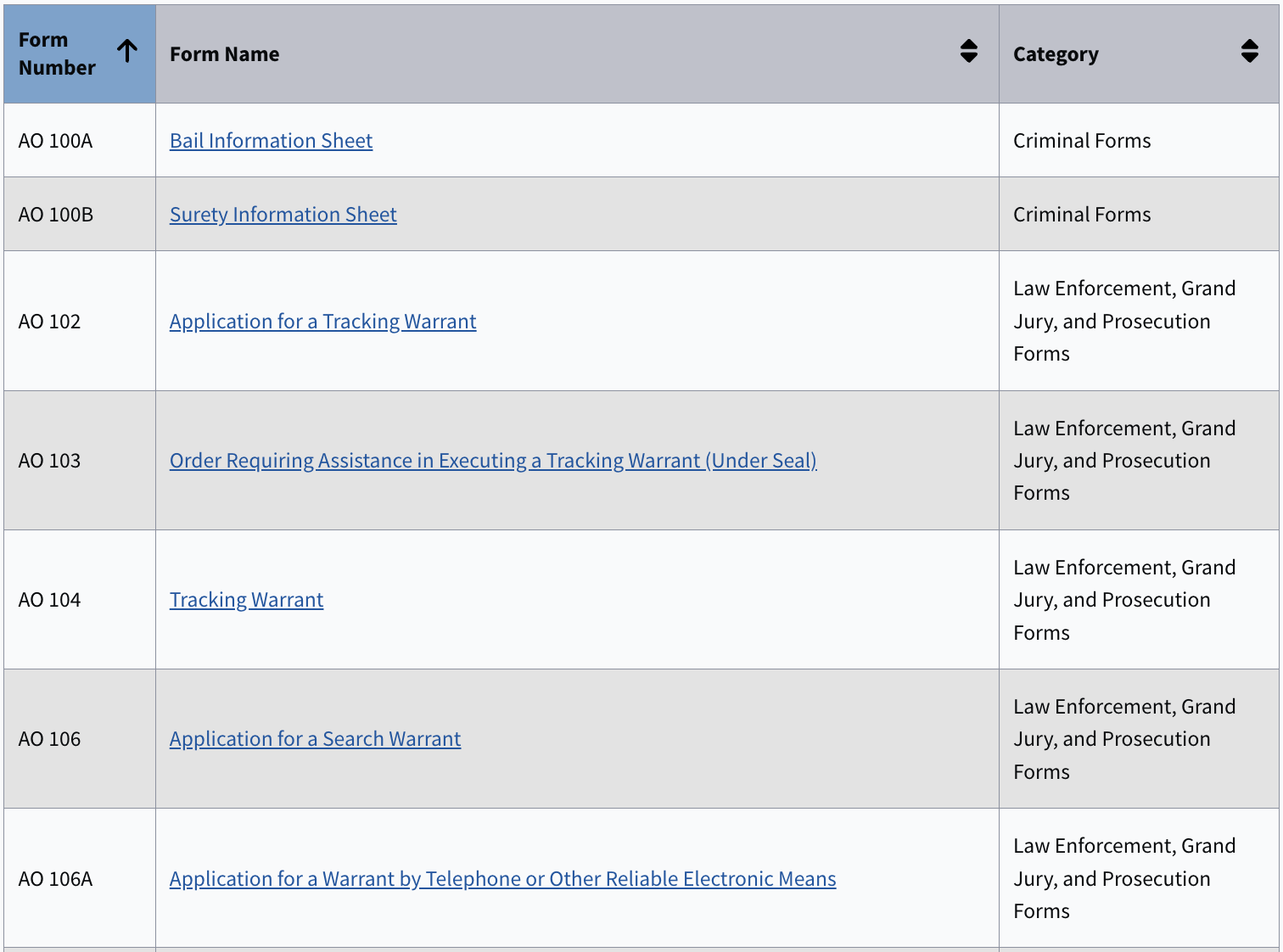} %
            \caption{} 
        \label{fig:horizontaltable}
        \end{subfigure}
        \begin{subfigure}{\linewidth}%
            \includegraphics[width=\linewidth]{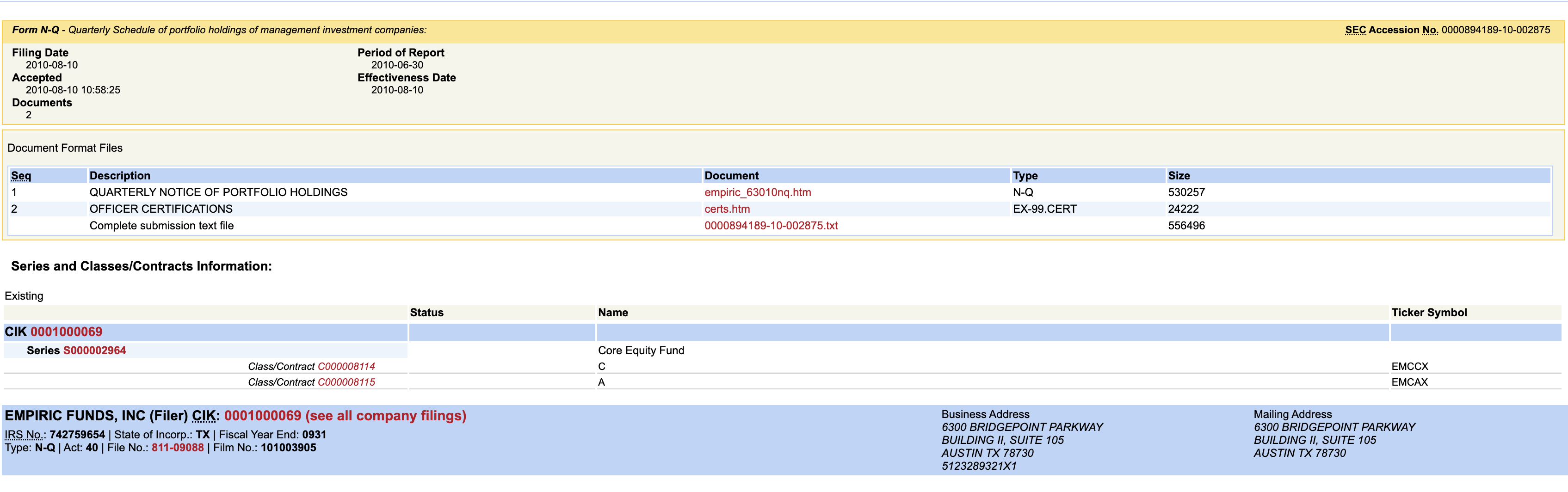} %
            \caption{} 
        \label{fig:freeform}
        \end{subfigure}
    \end{minipage}
    \begin{minipage}{0.325\textwidth}%
        \begin{subtable}{\linewidth}%
          \centering
          \small
            \begin{tabular}{p{0.18\linewidth}p{0.23\linewidth}p{0.39\linewidth}}
            \toprule
            \bf Subject & \bf Predicate & \bf Object \\
            \midrule
            AO 100A & Form Name & Bail Information Sheet \\
            \midrule
            AO 100A & Category & Criminal Forms \\
            \midrule
            AO 100B	& Form Name & Surety Information Sheet \\
            \midrule
            AO 100B & Category & Criminal Forms \\
            \midrule
            AO 102 & Form Name & Application for a Tracking Warrant	\\
            \midrule
            AO 102 & Category & Law Enforcement, Grand Jury, and Prosecution Forms \\
            \midrule
            AO 103 & Form Name & Order Requiring Assistance in Executing a Tracking Warrant (Under Seal) \\
            \midrule
            AO 103 & Category & Law Enforcement, Grand Jury, and Prosecution Forms \\
            \midrule
            AO 104 & Form Name & Tracking Warrant \\
            \midrule
            AO 104 & Category & Law Enforcement, Grand Jury, and Prosecution Forms \\
            \multicolumn{3}{c}{\bf $\cdots$} \\
            \bottomrule
            \end{tabular}
          \caption{}
          \label{tab:tripleannotation}
        \end{subtable}
    \end{minipage}%
    \caption{We categorize layouts of semi-structured data into three types: attribute-value pairs (\eg, Figure~\ref{fig:attributevalue}), horizontal tables (\eg, Figure~\ref{fig:horizontaltable}), free-form (\eg, Figure~\ref{fig:freeform}).  Figure~\ref{tab:tripleannotation} lists the extracted triples from Figure~\ref{fig:horizontaltable}.} 
    \label{fig:annotation}
\end{figure*}

\begin{figure*}[t!]
\centering
\includegraphics[width=0.9\textwidth]{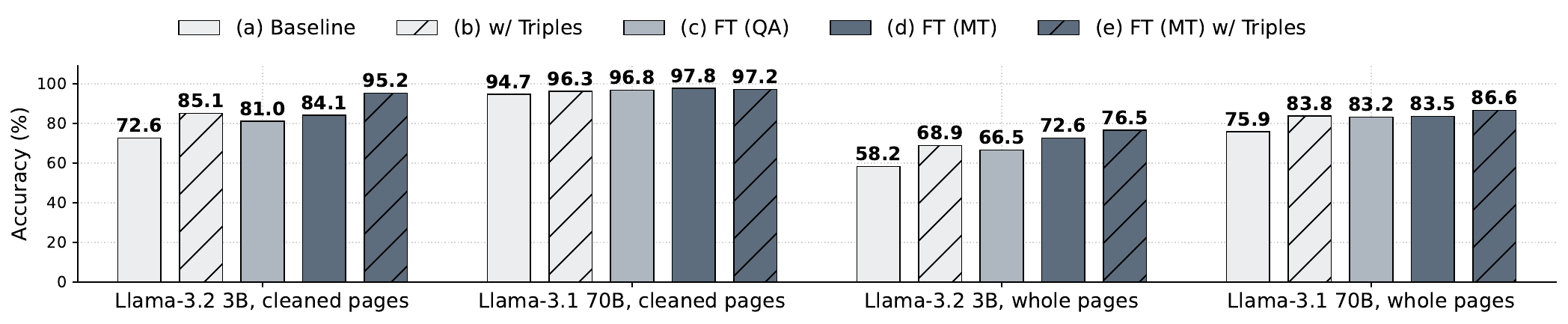}
\caption{Enhancing semi-structured content with ground truth knowledge triples during inference (\ie, (b)) surpasses the baseline without augmentation (\ie, (a)). Multi-task fine-tuning with triple extraction (d) outperforms QA-only fine-tuning  (\ie, (c)). Their combination (\ie, (e)) offers the greatest potential for improving QA performance in smaller models. (MT: multi-task, FT: fine-tuned.)}
\label{fig:intro}
\end{figure*}

In this paper, we investigate a specific category of web content: {\em semi-structured} webpages, where information is presented with some underlying structure such as web tables or attribute–value pairs (Figure~\ref{fig:annotation}), rather than in natural language texts. Our focus on semi-structured webpages is motivated by two key factors. First, the web contains a wealth of semi-structured data, typically populated by data from large underlying databases, thus offering a valuable source of factual knowledge~\cite{dong2014knowledge}. %
Second, it remains uncertain whether LLMs are equally adept at handling semi-structured content, which is rich in format but often lacks the grammatical signals to guide language understanding.

Knowledge extraction from semi-structured webpages has been extensively studied in the literature~\cite{lockard-etal-2020-zeroshotceres, lockard2019openceres, lockard2018ceres, Gulhane2011vertex, Kushmerick1997wrapper}. These studies highlight the unique challenges and opportunities posed by semi-structured content, %
differentiating it from extraction on unstructured texts~\cite{lockard2018ceres}. Despite the demonstrated high quality for the ClosedIE setting, where attributes are pre-defined and training examples abound, OpenIE, where attributes and even domains are unknown, remains difficult, with performance capped at a 46\% F-measure~\cite{lockard-etal-2020-zeroshotceres}.
This paper investigates three key questions to assess the value of knowledge extraction for QA on semi-structured data.

\smallskip
\par{\noindent \textbf{RQ1:}} How well do state-of-the-art LLMs extract knowledge from semi-structured web data?

\smallskip
\par{\noindent\textbf{RQ2:}} Do the extracted knowledge triples, added upon original semi-structured data, improve answer accuracy?

\smallskip
\par{\noindent\textbf{RQ3:}} Can equipping LLMs with explicit knowledge extraction capabilities enhance their QA performance?

\smallskip
Our foremost contribution towards addressing these questions is an extension of the WebSRC benchmark~\cite{chen-etal-2021-websrc} with more realistic and challenging data\footnote{The data will be available at \url{https://github.com/facebookresearch/SemiBench}.}. WebSRC was originally designed for QA over cleaned semi-structured content. We enhance it by incorporating original whole webpages from additional websites, similar to those encountered in RAG tasks with web retrieval. %
Additionally, we introduce a knowledge extraction task by annotating ground-truth knowledge triples, to enable direct evaluation of extraction quality and facilitate analysis of how extracted knowledge impacts QA performance. (Section~\ref{sec:data}-\ref{sec:setup}) %

We conducted comprehensive experiments using state-of-the-art LLMs for knowledge extraction (\textbf{RQ1}). Our findings reveal that LLMs obtain F1 scores of 77\% for triple extraction on cleaned semi-structured content, %
much higher than previous methods using GNN-based extraction (reported 46\% F-measure~\cite{lockard-etal-2020-zeroshotceres}). However, the score drops to 28\% on whole webpages, %
showing that {\em web-scale knowledge extraction remains challenging even with LLMs}. (Section~\ref{sec:extraction})

Our third contribution studies whether we can improve QA by augmenting the original semi-structured content with the extracted triples (\textbf{RQ2}). Our experiments show that although LLMs achieved high QA quality (95\%) on cleaned semi-structured content, they {\em struggle on whole webpages as appearing in real RAG systems,} and the quality drops to 76\%. Encouragingly, {\em augmenting the pages with ground-truth knowledge triples can increase QA accuracy} to 84\%; despite the potential, we cannot take for granted the availability of ground truth triples in practice. (Section~\ref{sec:rq3})

As a fourth contribution, we investigate whether improved capability of knowledge extraction can enhance LLMs' QA performance. We show that {\em multi-task fine-tuning on both QA and triple extraction tasks leads to improved QA performance even on unseen websites,} and the quality improvement is comparable to augmenting webpages with ground truth knowledge triples (83.5\% \vs 83.8\%). We also show that fine-tuning on triple extraction and providing extracted triples at inference time as augmentations offer \textit{complementary benefits}, and adding them together can further improve QA quality to 87\% (Figure~\ref{fig:intro}). (Section~\ref{sec:qa}).

Finally, as LLM-based knowledge extraction can be expensive, we investigate a new mechanism for knowledge extraction: we prompt LLMs with a few extraction examples, and let it write extraction scripts on semi-structured data. Remarkably, such methods can generalize across websites, yielding a 61\% F1-score on cleaned semi-structured content, comparable to fine-tuned LLMs (51\% (3B); 72\% (70B)). Additionally, the extracted triples can bring a 5\% gain in accuracy over zero-shot LLM-based QA---demonstrating \textit{a promising pathway for scalable knowledge extraction}. Despite the potential, we have not seen success on whole webpages, where scripting is much harder.

Our studies suggested that \textit{knowledge extraction can still effectively enhance LLMs' QA over semi-structured data}, through teaching of extraction capabilities and augmentation with extraction results. In addition to QA, knowledge extraction has broader values, such as offline indexing, knowledge integration, and user-facing presentation for improved interpretability (compared to unstructured content). Yet, \textit{the current quality of extraction at web scale remains sub-optimal}, and LLM-based extraction is expensive especially at the web scale. Improving it, especially with the low-cost script-based approach, is essential for enabling high-stakes, large-scale applications.

\section{Datasets} %
\label{sec:data}

\subsection{Problem definition}

Semi-structured data is a type of data that has some level of organization, but does not conform to the rigid structure of traditional relational databases. There are three forms of semi-structured data.

\par{\noindent\textbf{Attribute-value pairs / vertical tables (A-V):}} The data are in the form of attribute-value pairs (Figure~\ref{fig:attributevalue}). They may be in a table form, but with only two columns: one column for attributes, and one column for values.

\par{\noindent\textbf{Horizontal web tables (Hz):}} The table typically has multiple rows and columns (Figure~\ref{fig:horizontaltable}). Normally a row represents an entity and a column represents an attribute, and the corresponding cell gives the value. There are also cases where a column represents an entity and a row represents an attribute.

\par{\noindent\textbf{Free form semi-structured data (F-F):}} Free forms typically has a subject on the top of the page or a chunk, and key-value attributes in various layouts, horizontally aligned or vertically aligned (Figure~\ref{fig:freeform}); sometimes the predicates are skipped. %

We consider the two problems below on semi-structured data. %

\par{\noindent\textbf{Question answering.}} Given a question and an associated semi-structured reference document as input, the goal is to output the answer.
\par{\noindent\textbf{Triple extraction.}} Given a semi-structured document as input, the goal is to generate a structured representation in text, typically in the form of (subject, predicate, object) triples.

\begin{table}[t]
\centering
\small

\caption{Statistics of the enriched dataset.}
\begin{tabular}{lr}
\toprule
 \bf Metric & \bf Value \\
\midrule
\textbf{\textit{Cleaned pages}} \\
\# QA pairs / \# webpages  & 6,186 / 180 \\
\ \ $\diamond$ A-V & 1,089 / 72 \\ 
\ \ $\diamond$ Hz & 3,224 / 51 \\
\ \ $\diamond$ F-F & 1,873 / 57  \\
 \# triples & 4,741 \\
 \# triples per page (average) & 26.3 \\
\midrule
\textbf{\textit{Whole pages}} \\
\# QA pairs / \# webpages  & 1,006 / 251\\
\# triples & 30,263 \\
\# triples per page (average) & 120.6 \\
\bottomrule
\end{tabular}
\vspace{-.1in}
\label{tab:datastat}
\end{table}

\subsection{The enriched WebSRC benchmark}

\par{\noindent\textbf{Cleaned webpages.}} Part of our work is built upon WebSRC, the only large-scale semi-structured QA dataset available. Comprising question-answer pairs from 70 diverse websites across various topics and layouts, this dataset offers a comprehensive representation of the complexities inherent in semi-structured data. Webpages in this dataset have been %
cleaned, with noise elements such as ads and navigation tabs removed, and large objects truncated~\cite{chen-etal-2021-websrc}. Among these 70 websites, 60 have publicly accessible ground truth QA annotations. We focus on these 60 websites and extend them with 
a new task -- triple extraction: we extract all (subject, predicate, object) triples from the HTML (HyperText Markup Language) of a webpage, in the order they appear on the page. %

Specifically, we randomly selected three webpages from each of the 60 websites %
and manually annotated all triples present in these webpages. The subjects, predicates, and objects generally adhere to the original wording, primarily derived from text spans within the webpage. However, annotators were permitted to add supplementary disambiguation information when necessary; such information had to be enclosed in special characters. They may skip any incomplete triples (\eg, missing predicate) if a part is missing from the webpage. An example is shown in Figure~\ref{fig:horizontaltable} and \ref{tab:tripleannotation}, %
and see Appendix~\ref{sec:annotationdetails} for more details. %

To facilitate the investigation of the connection between triple extraction and question answering, we leveraged QA pairs from WebSRC that correspond to the same set of sampled webpages described above. Given the presence of paraphrased questions in WebSRC, we only retained one instance of each question with a unique ground truth answer. Table~\ref{tab:datastat} summarizes the overall statistics.

\par{\noindent\textbf{Whole webpages.}} Additionally, we sampled 139 websites with semi-structured content from the top 2,500 registered domains ranked by page captures in Common Crawl\footnote{\url{https://commoncrawl.org/}.}. For 56 of these sampled websites, we collected three webpages each; for the remaining 83 websites, we collected one webpage each. We manually annotated all triples on these webpages using the same process as for cleaned webpages, and developed human-audited QA pairs for each webpage with the support of LLMs. See Table~\ref{tab:datastat} for statistics and Appendix~\ref{sec:fullpageqapairgen} for more details. 

\subsection{Evaluation metrics}
\par{\noindent\textbf{QA accuracy.}} We use \textbf{Accuracy\textsubscript{LM}}, defined as the percentage of correctly answered questions, where we utilize an LLM (Llama 3.1-70B-Instruct) to check whether the response aligns with the ground truth. Llama was selected for its open-source nature and reproducibility. We verified human judgment against this metric for 150 random instances, with over 99\% agreement. An additional rule-based metric is reported and discussed in Appendix~\ref{sec:supplementalresults}.

\par{\noindent\textbf{Triple extraction quality.}} We evaluate the performance of triple extraction using both global and triple-level matching metrics. 

\smallskip
\par{\noindent \textbf{-~Global match.}} We use \textbf{fuzzy match (FM)}, defined as the normalized character-level edit distance between the prediction and the ground truth, considering all extracted triples as a whole. 

\smallskip
\par{\noindent \textbf{-~Triple-level match.}} We first calculate the character-level edit distance between each pair of triples from the prediction and the ground truth, resulting in a score matrix. We then obtain the maximum weight matching of the score matrix using the Munkres algorithm~\cite{munkres1957algorithms}. Next, we employ an LLM (Llama 3.1-70B-Instruct) to check whether each pair of triple from the maximum weight matching are semantically the same. Based on this process, we define the following metrics:
\begin{itemize}
\item \textbf{LLM-based Precision (P\textsubscript{LM}):} The percentage of extracted triples that are matched to ground-truth triples.
\item \textbf{LLM-based Recall (R\textsubscript{LM}):} The percentage of ground-truth triples that are matched to extracted triples.
\item \textbf{LLM-based F-1 Score (F-1\textsubscript{LM}):} The harmonic mean of precision and recall, calculated as  $2\cdot\text{P\textsubscript{LM}}\cdot\text{R\textsubscript{LM}}/(\text{P\textsubscript{LM}}+\text{R\textsubscript{LM}})$ 
\end{itemize}
We manually verified the agreement between human judgment and the LLM for 150 randomly sampled triple pairs. The results showed a 95\% agreement rate, which we consider acceptable, given that some cases may be open to interpretation. We additionally report and discuss other triple-level metrics in Appendix~\ref{sec:supplementalresults}.

During evaluation, we disregard disambiguation information and incomplete triples in both the prediction and the ground truth.

\section{Experimental Setup}
\label{sec:setup}

\subsection{Configurations}

For cleaned webpages, we employ two distinct configurations when utilizing the datasets:

\par{\noindent \textbf{In-domain.}} We split the data such that we use the same set of websites during training and evaluation, with 120 webpages (two webpages per website) for training and 60 webpages (a webpage per website) for evaluation. 

\par{\noindent\textbf{Out-of-domain.}} We split the data such that there is minimal domain overlap between the training and evaluation websites, with 90 webpages for training and 90 for evaluation.

For whole webpages, among the 56 websites from which we collected three webpages each, we used two webpages per website (a total of 112 webpages) for model training. The remaining webpages from these 56 websites were used for  \textbf{in-domain} evaluation. Additionally, the 83 websites from which only one webpage was collected were used for \textbf{out-of-domain} evaluation.

\subsection{Foundational models}

We primarily focus on the experimental results obtained using Llama 3.2-3B-Instruct (L-3B) and Llama 3.1-70B-Instruct (L-70B)~\cite{grattafiori2024llama3herdmodels}. For a more comprehensive understanding, we also report the performance of Qwen 2.5-3B-Instruct (Q-3B), Qwen 2.5-72B-Instruct (Q-72B)~\cite{qwen2025qwen25technicalreport}, GPT-4o (4o), GPT-4o mini (4o mini)~\cite{openai2024gpt4ocard}, and Claude 3.7 Sonnet\footnote{\url{https://www.anthropic.com/news/claude-3-7-sonnet}} where relevant. Training and inference details are available in Appendix~\ref{sec:implementationdetail}, and prompt details are available in Appendix~\ref{sec:prompt}.

\section{Triple Extraction (RQ1)}
\label{sec:extraction}

\subsection{Experiment setup}
We considered three types of approaches for triple extraction:

\smallskip
\par{\noindent \bf (i) Off-the-shelf LLM-based extractors.} 
\par{\noindent \bf In-domain:} We employ zero-shot and 2-shot extraction (Prompt~\ref{prompt:tezeroshot} and \ref{prompt:tefewshot} in Appendix~\ref{sec:prompt}). For the 2-shot setting, each example contains a webpage from the same website, demonstrating triples that shall be extracted from the webpage. 
\par{\noindent \bf Out-of-domain:} We use zero-shot or 3-shot extraction, where the latter provides three examples each from the three forms of semi-structured content (\ie, attribute-value pair, horizontal table, and free-form). The out-of-domain setting illustrates web-scale extraction, where even providing two examples per website is still impractical.

\smallskip
\par{\noindent \bf (ii) Fine-tuned LLM-based extractors.} We fine-tune the LLM using in-domain and out-of-domain training sets for respective configurations. Each training instance contains a webpage and the ground truth triples embraced by the page. 

\smallskip
\par{\noindent \bf (iii) Triple extraction through automatically generated scripts.} Running LLMs on each webpage can be prohibitively expensive. We employ Llama 3.1-70B-Instruct to generate a script for triple extraction from HTML for each website, and then run the scripts for triple extraction, which is much less resource-consuming.
\par{\noindent \bf Single call:} We provide the LLM with two exemplar instances of triple extraction, prompting it to generate a script. For in-domain extraction, the exemplar instances consist of HTML pages from the same website in the training set, along with their corresponding ground truth triples. For out-of-domain extraction, the exemplar instances consist of HTML pages from other webpages of the same website in the evaluation set, paired with triples extracted by the 3-shot extraction model, since ground truth triples are unavailable.
\par{\noindent \bf Multiple calls with feedback:} We generate multiple scripts through multiple calls and iteratively refine the generated scripts by executing the LLM-generated script and providing the execution results as feedback to the LLM. We select the best one based on a chosen metric evaluated on the exemplar instances. See Appendix~\ref{sec:ablationscript} for more details.

\begin{table*}[ht!]
\small
\centering
\caption{Triple extraction performance on cleaned pages. All numbers are in percentage (\%).}
\begin{tabular}{lllrrrr}
\toprule
 & \bf Backbone & \bf Setting & \multicolumn{1}{c}{\bf Global} & \multicolumn{3}{c}{\bf Triple-level} \\
& & & FM & P\textsubscript{LM} & R\textsubscript{LM} & F-1\textsubscript{LM} \\
\cmidrule(lr){1-4} \cmidrule(lr){5-7}

\multirow{10}{*}{in-domain} 
 & Llama 3.2-3B-Instruct & zero-shot & 49.6 & 31.7 & 45.9 & 37.5  \\ %
 & Llama 3.2-3B-Instruct & 2-shot & 48.7 & 50.4 & 73.6 & 59.8 \\ %
 & Llama 3.2-3B-Instruct & fine-tuned & 69.3 & 60.1 & 63.7 & 61.9 \\ 
\cmidrule(lr){2-4} \cmidrule(lr){5-7} 

& Llama 3.1-70B-Instruct & zero-shot & 57.8 &  53.1 & 71.8 & 61.0  \\ %
& Llama 3.1-70B-Instruct & 2-shot & 90.2 & 88.4 & 92.0 & 90.2 \\ %
& Llama 3.1-70B-Instruct & fine-tuned & 78.5 & 71.8 & 76.3 & 74.0 \\ %
 & Claude 3.7 Sonnet & 2-shot & 92.4 & 89.9 & 92.8 & 91.3 \\ %
 & GPT-4o & 2-shot & 94.1 & 93.9 & 95.5 & 94.7 \\ %
\cmidrule(lr){2-4} \cmidrule(lr){5-7} 

 & generated scripts & single call & 59.0 & 53.1 & 49.3 & 51.1 \\ %
 & generated scripts & multiple calls with feedback & 74.5 & 75.8 & 69.9 & 72.7  \\ %
\cmidrule(lr){1-4} \cmidrule(lr){5-7} 

\multirow{10}{*}{out-of-domain} 
& Llama 3.2-3B-Instruct & zero-shot & 49.7 & 31.3 & 42.3 & 36.0 \\ %
& Llama 3.2-3B-Instruct & 3-shot & 45.7 & 26.5 & 29.2 & 27.8  \\ %
& Llama 3.2-3B-Instruct & fine-tuned & 64.0 & 52.0 & 50.6 & 51.3   \\ %
\cmidrule(lr){2-4} \cmidrule(lr){5-7}

& Llama 3.1-70B-Instruct & zero-shot & 58.6 & 60.3 & 74.1 & 66.5  \\ %
& Llama 3.1-70B-Instruct & 3-shot & 74.1 & 69.4 & 70.1 & 69.7  \\  %
& Llama 3.1-70B-Instruct & fine-tuned & 75.7 & 72.1 & 71.0 & 71.5 \\ %
& Claude 3.7 Sonnet & 3-shot & 78.3 & 76.0 & 77.4 & 76.7 \\ %
& GPT-4o & 3-shot & 75.9 & 76.4 & 76.9 & 76.6  \\ %

\cmidrule(lr){2-4} \cmidrule(lr){5-7}
 & generated scripts & single call & 59.0 & 50.9 & 45.6 & 48.1 \\
 & generated scripts & multiple calls with feedback & 64.4 & 63.8 & 57.9 & 60.7 \\
\bottomrule
\end{tabular}

\label{tab:tripleextractionall}
\end{table*}

\begin{table*}[ht!]
\small
\centering
\caption{Triple extraction performance on whole webpages. All numbers are in percentage (\%).}
\begin{tabular}{llllrrrr}
\toprule
 & \bf Backbone & \bf Page Cleaning & \bf Setting & \multicolumn{1}{c}{\bf Global} & \multicolumn{3}{c}{\bf Triple-level} \\
& & & & FM & P\textsubscript{LM} & R\textsubscript{LM} & F-1\textsubscript{LM} \\
\cmidrule(lr){1-5} \cmidrule(lr){6-8}

\multirow{3}{*}{in-domain} & Llama 3.1-70B-Instruct & / & zero-shot & 17.4 & 17.8 & 11.0 & 13.6  \\
 & Llama 3.1-70B-Instruct & Trafilatura & zero-shot & 30.1 & 22.8 & 12.5 & 16.1 \\
& Llama 3.1-70B-Instruct & Trafilatura & fine-tuned & 34.4 & 22.8 & 15.1 & 18.2 \\ 
\cmidrule(lr){1-5} \cmidrule(lr){6-8} 

\multirow{5}{*}{out-of-domain} & Llama 3.1-70B-Instruct & / & zero-shot & 17.6 & 12.5  & 12.6  & 12.5  \\
& Llama 3.1-70B-Instruct & Trafilatura & zero-shot & 33.3 & 24.3 & 15.7 & 19.1 \\
& Llama 3.1-70B-Instruct & Trafilatura & fine-tuned & 39.5 & 27.1 & 21.4 & 23.9 \\
\cmidrule(lr){2-5} \cmidrule(lr){6-8} 
 & Claude 3.7 Sonnet & Trafilatura & zero-shot & 32.7 & 26.2 & 23.0 & 24.5 \\
 & GPT-4o & Trafilatura & zero-shot & 30.9 & 35.1 & 23.8 & 28.3 \\ 
\bottomrule
\end{tabular}

\label{tab:fullpagetripleextractionall}
\end{table*}

\subsection{Triple extraction quality}
{\par{\noindent \bf Cleaned pages.}} We present in Table~\ref{tab:tripleextractionall} the overall performance of various triple extraction approaches on cleaned pages. 

\paragraph{In-domain extraction.} We have the following four observations. (1) Zero-shot triple extraction has mediocre performance on semi-structured data, obtaining triple-level F-1 of 61\% with Llama 3.1-70B. (2) With only two examples from the same website, the 70B model quality (F-1) improved significantly to 90\%, meaning as far as we manually annotate a couple of pages, we can achieve reasonable extraction quality. (3) The fine-tuned 70B model underperforms 2-shot, likely because the training data contains a broader set of 120 webpages from 60 websites, losing focus on the specific website during inference compared to 2-shot in-content learning. (4) Notably, LLM-based script-writing is almost comparable to the fine-tuned 70B model, with much lower computation resources for \emph{inference}-time extractions. The inference time of the generated scripts on a single Xeon Platinum 8339HC core is approximately 1/3000th of the inference time of a 3B LLM running on a single A100 (40GB) GPU. However, note that script coding poses considerable difficulties compared to directly understanding and converting semi-structured contents to triples; the multiple-call approach provides significant improvement over single-call for script generation (further ablation study in Appendix~\ref{sec:ablationscript}). 

\paragraph{Out-of-domain extraction.} In contrast, out-of-domain triple extraction is still challenging. We highlight three differences from in-domain extraction. First, compared to zero-shot, 3-shot in-context learning does not help much, since different websites often have very different formats. Second, fine-tuning helps better than 3-shot in-context learning; the improvement is marginal on 70B model (+2\% on F1), but much more pronounced on 3B model (+24\%). Finally, overall quality is much lower than in-domain---all state-of-the-art models have F1-score below 80\%, showing that triple extraction is not ready for web scale even with LLMs. 

\paragraph{Triple extraction on different forms.} Table~\ref{tab:tripleextractionlayout} in Appendix summarizes the performance across the three different layouts. We observed the highest quality for the key-value form: even for out-of-domain, we achieved up to 90\% F-1. In contrast, the quality for horizontal tables is lower (up to 81\% F-1), and for free-form is the lowest (up to 63\%). This is as expected, as there is the least commonality among websites for free-form data, making out-of-domain extraction more challenging. %

\smallskip
{\par{\noindent \bf Whole pages.}} We present in Table~\ref{tab:fullpagetripleextractionall} the performance on whole webpages. Note that the context window of popular LLMs, including those we benchmarked, is typically limited to 128K tokens, which poses challenges for processing whole webpages. Specifically, the average and maximum token counts in our evaluation set (measured using the tokenizer of Llama 3.1) are 110K and 1.1M tokens, respectively. As a result, 29\% of whole webpages exceed the 128K context window and cannot be processed. For these instances, we assign their quality metrics (\ie, P\textsubscript{LM}, R\textsubscript{LM}, etc.) a value of zero. To mitigate this issue, we apply Trafilatura~\cite{barbaresi-2021-trafilatura} for automatic page cleaning, which removes non-content noise (\eg, headers, footers, ads).  This significantly reduces the average and maximum token counts to 6K and 105K, respectively. We did not benchmark few-shot extraction and LLM-based script writing performance because the maximum sequence length they require exceeds 128K tokens, even after page cleaning. We highlight two observations: First, the extraction quality is substantially lower, with the highest F-1\textsubscript{LM} reaching only 28\%, indicating a significant gap between triple extraction from cleaned pages versus whole pages. Second, page cleaning generally contributes to improved triple extraction performance.

\smallskip
{\par{\noindent \bf Summary:}} Even with the advanced understanding capabilities of LLMs, triple extraction remains a challenge on semi-structured content. Extraction on cleaned semi-structured content achieves up to 77\% F1-score on out-of-domain websites; on full webpages, the F1-score drops to 28\%, showing the difficulties in web-scale extraction. Encouragingly, we observe that script-based extraction achieves comparable results to few-shot extraction, offering an alternative low-cost medium-quality approach.

\begin{table}[t!]
\small
\centering
\caption{Question answering performance on whole webpages in Accuracy\textsubscript{LM} (\%).}
\begin{tabular}{lllr}
\toprule
& \bf Backbone & \bf Cleaning & \bf Accuracy\textsubscript{LM} \\
\cmidrule(lr){1-4}
\multirow{3}{*}{in-domain} & Llama 3.1-70B-Instruct  & Trafilatura & 49.3 \\ 
& Llama 3.1-70B-Instruct  & / & 76.2 \\ 
\cmidrule(lr){1-4}
\multirow{5}{*}{out-of-domain} & Llama 3.1-70B-Instruct & Trafilatura & 51.8 \\ %
& Llama 3.1-70B-Instruct & /  & 75.9 \\  %
\cmidrule(lr){2-4}
& GPT-4o & / & 86.3 \\ 
& Claude 3.7 Sonnet & / & 77.1 \\

\bottomrule
\end{tabular}

\label{tab:fullpagequestionansweringall}
\end{table}

\begin{table*}[t!]
\small
\centering
\caption{Out-of-domain QA performance in Accuracy\textsubscript{LM} (\%). } %
\begin{tabular}{llrrrrrr}
\toprule
& \bf Augmentation & \bf L-3B & \bf L-70B & \bf Q-3B & \bf Q-72B & \bf 4o mini & \bf 4o \\
\cmidrule(lr){1-4} \cmidrule(lr){5-6} \cmidrule(lr){7-8} 
\multirow{2}{*}{cleaned pages} & no augmentation & 72.6 & 94.7 & 81.6 & 95.1 & 93.3 & 96.3 \\ %
& with ground truth triples & \bf 85.1 & \bf 96.3 & \bf 92.4 & \bf 97.1 & \bf  97.0 & \bf 96.7 \\ %
\cmidrule(lr){1-4} \cmidrule(lr){5-6} \cmidrule(lr){7-8} 
\multirow{2}{*}{whole pages} & no augmentation & 58.2 & 75.9 & 61.0  & 81.4 & 77.1 & 86.3  \\ 
& with ground truth triples & \bf 68.9 & \bf 83.8 & \bf 72.3 & \bf 85.1 & \bf 84.1 & \bf 90.5 \\
\bottomrule
\end{tabular}

\label{tab:additionalref}
\end{table*}

\begin{table*}[t!]
\centering
\small
\caption{Zero-shot QA performance in Accuracy\textsubscript{LM} (\%) on cleaned pages. Script-extracted triples improve QA quality for 3B models. }
\begin{tabular}{llrrrrr}
\toprule
 \bf  & \bf Additional reference & \bf L-3B & \bf L-70B & \bf Q-3B & \bf Q-72B \\
\midrule
\multirow{2}{*}{in-domain} & / & 77.1 & 95.1 & 81.5 & 93.5 \\ %
& Script-extracted triples & 80.6 & 94.9 & 87.5 & 94.2   \\ %
\cmidrule(lr){1-6}
\multirow{2}{*}{out-of-domain} & / &  72.6 & 94.7 & 81.6 & 95.1   \\ 
& Script-extracted triples & 77.7 & 92.9 & 86.5 & 94.2  \\ %

\bottomrule
\end{tabular}

\label{tab:indomainqa}
\end{table*}

\section{Improving QA Through Triple Augmentation (RQ2)}
\label{sec:rq3}
We next examine LLMs' quality in question answering leveraging semi-structured content, and whether or not augmenting the original content with the extracted knowledge triples can enhance QA performance. For this purpose, we conducted experiments where we concatenate the extracted triples obtained from a triple extraction model with the original reference text to create a new reference. %

\paragraph{QA on semi-structured data.} Table~\ref{tab:additionalref} reports results for out-of-domain QA. Whereas smaller models (\ie, Llama 3B, Qwen 3B, and GPT-4o mini) obtain only medium QA accuracy on cleaned pages, larger models (\ie, Llama 70B, Qwen 72B, and GPT-4o) already achieve high accuracy (94-96\%). However, the quality drops significantly on whole webpages; even larger models achieve at best 86\% QA accuracy. The nuance is that we use the flattened page for QA (see Appendix~\ref{sec:implementationdetail} for details), which is shorter than the original HTML. Page cleaning using Trafilatura reduces the page length but at the cost of mistakenly missing a substantial amount of useful information (49.3\% \vs 76.2\%), highlighting the gap between information extraction from cleaned pages versus whole pages using LLMs (see Table~\ref{tab:fullpagequestionansweringall}).

\paragraph{Ground-truth triples.} We start with ground truth triples for augmentation to illustrate the upper-bound benefit. Table~\ref{tab:additionalref} shows that incorporating ground truth triples can significantly enhance the QA performance: for smaller models, which have only medium quality on cleaned pages, the augmentation can improve by up to 13\%; for  larger models, which still fall short on whole pages, ground-truth triples bring substantial benefits and increase accuracy by up to 8\%. Despite the potential benefits, ground truth triples are not readily available for augmentation.

\paragraph{Script-extracted triples.} We further experiment with augmentation using script-extracted triples, which are more feasible in practice but may contain a level of noises. Table~\ref{tab:indomainqa} presents overall QA quality. Even when augmented with script-extracted triples, we already observe much higher QA accuracy for 3B models for zero-shot QA, improving accuracy by up to 6\%. Further analysis reveals that this trend also holds for 2-shot QA, and that the primary source of improvement comes from better performance on horizontal tables (Table~\ref{tab:qascriptextractedtripleslayout} in Appendix). %
Larger models already obtain high QA accuracy on cleaned pages, thus do \textit{not} gain additional benefits from script-based extractions because of the potential distractions caused by wrongly extracted triples. %

\smallskip
{\par{\noindent \bf Summary:}} Whereas LLMs already achieve high QA quality (96\%) on cleaned semi-structured content, the quality drops by up to 20\% on whole pages in the wild. Augmenting webpages with triples that capture factual knowledge from the semi-structured content can improve QA quality substantially. In practice, however, noisy triples such as those extracted by scripts, improve more for smaller LLMs but less for large LLMs that already achieve high QA quality.

\begin{table*}[ht!]
\small
\centering
\caption{Question answering performance in Accuracy\textsubscript{LM} (\%) on cleaned pages.}
\begin{tabular}{lllrrrr}
\toprule
& \bf Backbone & \bf Setting & \bf All & \bf A-V & \bf Hz & \bf F-F \\
\cmidrule(lr){1-7}
\multirow{2}{*}{in-domain} & Llama 3.2-3B-Instruct & zero-shot  & 77.1 & 85.9  & 72.4  & 80.0 \\ %
& Llama 3.2-3B-Instruct & fine-tuned  & 98.2 & 97.0 & 99.6 & 96.5  \\ %

\cmidrule(lr){1-7}
\multirow{6}{*}{out-of-domain} & Llama 3.2-3B-Instruct &  zero-shot & 72.6  & 81.1  & 70.1 & 81.8 \\ %
& Llama 3.2-3B-Instruct &  fine-tuned  & 81.0 & 97.5 & 76.7 & 96.0 \\ %
& Llama 3.1-70B-Instruct & zero-shot   & 94.7  & 91.8  & 94.7  & 96.8  \\ %
& Llama 3.1-70B-Instruct & fine-tuned  & 96.8 & 99.1 & 96.7 & 95.6 \\  %
\cmidrule(lr){2-7}
& GPT-4o & zero-shot  & 96.3 & 94.0 & 96.3 & 98.0 \\ %
& Claude 3.7 Sonnet & zero-shot  & 97.2 & 95.9 & 97.2 & 98.2 \\ %

\bottomrule
\end{tabular}

\label{tab:questionansweringall}
\end{table*}

\begin{table*}[t!]
\centering
\small
\caption{Out-of-domain QA performance in Accuracy\textsubscript{LM} (\%) across different layouts on cleaned pages. Compared to solely fine-tuning on the QA task, multi-task fine-tuning with triple extraction consistently improves results on horizontal (Hz) tables and free-form (F-F) data.}
\begin{tabular}{lrrrrrrrrrrrrr}
\toprule
\bf Fine-tuning task & \multicolumn{3}{c}{\bf L-3B} & \multicolumn{3}{c}{\bf L-70B} & \multicolumn{3}{c}{\bf Q-3B} & \multicolumn{3}{c}{\bf Q-72B} \\
 & A-V & Hz & F-F & A-V & Hz & F-F  & A-V & Hz & F-F  & A-V & Hz & F-F\\
\cmidrule(lr){1-4} \cmidrule(lr){5-7} \cmidrule(lr){8-10} \cmidrule(lr){11-13}
QA & 97.5 & 76.7 & 96.0 & 99.1 & 96.7 & 95.6 & 94.7 & 81.5 & 82.8 & \bf 99.1 & 96.5 & 96.0 \\  %
QA + Triple Extraction & \bf 97.8 & \bf 80.3 & \bf 97.8 & \bf 99.4 & \bf 97.5 & \bf 98.6 & \bf 97.2 & \bf 84.4 & \bf 84.2 & 96.5 & \bf 97.0 & \bf 98.6 \\ %
\bottomrule
\end{tabular}

\label{tab:multitaskqalayout}
\end{table*}

\begin{table*}[t!]
\small
\centering
\caption{Out-of-domain QA performance in Accuracy\textsubscript{LM} (\%) on cleaned pages. }
\begin{tabular}{llrrrr}
\toprule
\bf Fine-tuning task & \bf Additional reference & \bf L-3B & \bf L-70B & \bf Q-3B & \bf Q-72B \\
\cmidrule(lr){1-4} \cmidrule(lr){5-6}
\multirow{4}{*}{QA} & / & 81.0 & 96.8 & 82.8 & 96.7 \\  %
 & FT 3B extracted triples & 78.6 & / & 82.7  & /\\ %
 & FT 70+B extracted triples & 87.0 & 96.0 & 83.4 & 95.9 \\ %
 & Ground truth triples & 94.6 & 96.9 & 93.5  & 96.8 \\ %
\cmidrule(lr){1-4} \cmidrule(lr){5-6} 
\multirow{4}{*}{QA + Triple Extraction} & / & 84.1 & 97.8 & 85.4 & 97.2 \\ %
 & FT 3B extracted triples & 84.0 & / & 84.1 & /\\ %
& FT 70+B extracted triples &  89.4 & 96.6 & 85.4 & 96.3 \\ %
& Ground truth triples & 95.2 & 97.2 & 94.5 & 97.1 \\ %
\bottomrule
\end{tabular}

\label{tab:additionalref_finetuned}
\end{table*}

\begin{table*}[t!]
\small
\centering
\caption{QA performance on whole webpages from out-of-domain websites in Accuracy\textsubscript{LM} (\%). }
\begin{tabular}{llrrrr}
\toprule
\bf Fine-tuning task & \bf Additional reference & \bf L-3B & \bf L-70B & \bf Q-3B & \bf Q-72B \\
\cmidrule(lr){1-4} \cmidrule(lr){5-6}
\multirow{2}{*}{/} & / & 58.2 & 75.9 & 61.0 & 81.4 \\  
 & Ground truth triples & 68.9 & 83.8 & 72.3 & 85.1 \\
 \cmidrule(lr){1-4} \cmidrule(lr){5-6} 
\multirow{4}{*}{QA} & / & 66.5 & 83.2 & 66.5 & 82.6 \\  
 & FT 3B extracted triples & 69.2 & / & 67.7 & / \\
& FT 70+B extracted triples & 70.1 & 82.3 & 68.6 & 82.3 \\
 & Ground truth triples & 76.2 & 87.5 & 73.5 & 85.7 \\
\cmidrule(lr){1-4} \cmidrule(lr){5-6} 
\multirow{4}{*}{QA + Triple Extraction} & / & 72.6 & 83.5 & 68.0 & 82.0 \\
 & FT 3B extracted triples & 71.0 & / & 67.1 & / \\
& FT 70+B extracted triples & 70.1 & 84.5 & 71.6 & 82.3  \\
& Ground truth triples & 76.5 & 86.6 & 74.7 & 86.6 \\
\bottomrule
\end{tabular}

\label{tab:fullpageadditionalref_finetuned}
\end{table*}

\begin{table}[t!]
\centering
\small
\caption{Out-of-domain QA performance in Accuracy\textsubscript{LM} (\%) on cleaned pages. Multi-task fine-tuning (FT) with triple extraction (TE) enhances QA performance. }
\begin{tabular}{lrrrrr}
\toprule
\bf FT task & \bf L-3B & \bf L-70B  &  \bf Q-3B & \bf Q-72B \\
\midrule
/ & 72.6 & 94.7 & 81.6 & 95.1 \\ %
QA & 81.0 & 96.8 & 82.8 & 96.7 \\  %
TE & 74.3  & 94.6  & 81.4  & 95.3  \\ %
QA + TE & \bf 84.1 & \bf 97.8 & \bf 85.4 & \bf 97.2 \\ %
\bottomrule
\end{tabular}

\label{tab:multitaskqa}
\end{table}

\section{Improving QA Trough Enhanced Triple Extraction Capabilities (RQ3)}
\label{sec:qa}
We next examine whether by teaching LLMs to obtain triple-extraction capability can enhance QA on cleaned pages. We perform multi-task fine-tuning on both QA and triple extraction tasks. See Appendix~\ref{sec:implementationdetail} for implementation details. %

\smallskip
{\par{\noindent \bf Cleaned pages.}} 
We summarize the QA performance of zero-shot and fine-tuned LLMs in Table~\ref{tab:questionansweringall}. The fine-tuned Llama 3B achieves strong performance (98\%) on in-domain QA.

We have three observations on out-of-domain QA. First, it is harder than in-domain. State-of-the-art commercial models (GPT-4o and Claude 3.7 Sonnet) and Llama 70B all demonstrate strong zero-shot performance (up to 97\% in accuracy) in question answering. %
Second, fine-tuning lifts QA accuracy, significantly for the 3B model and marginally for the 70B model. Third, fine-tuning normally helps when original quality is sub-optimal, such as on attribute-value pairs for the 70B model, and on free-form for the 3B model; however, even after fine-tuning, a major gap remains on horizontal tables for 3B model.

\paragraph{Effect of multi-task training.} We next investigate if the knowledge-extraction capability improves QA capability. Table~\ref{tab:multitaskqa} compares the performance of LLMs fine-tuned on different tasks. Multi-task fine-tuning with triple extraction provides additional benefits, particularly for 3B models. 
Moreover, as shown in Table~\ref{tab:multitaskqalayout}, multi-task fine-tuning primarily and consistently improves QA accuracy on the two harder forms: horizontal tables and free-form data, over sole fine-tuning on the QA task.

\paragraph{Adding in addition triple augmentation.} Finally, we augment the input with extracted triples and run the fine-tuned models. Table~\ref{tab:additionalref_finetuned} reports results for out-of-domain QA. %
We observe that fine-tuning is not as effective as triple augmentation.  For instance, when Llama 3B is fine-tuned for the QA task, adding multi-task fine-tuning yields a 3\% improvement in quality, whereas incorporating triples extracted by 70+B models boosts quality by 6\%. 
With that being said, these two approaches are complementary to each other. Continuing with the same example, combining the two improves QA accuracy by 8\%. Despite the combined power, we point out that the QA models generally do not benefit from triples extracted by fine-tuned models of the same size, which do not seem to add additional value.

\smallskip
{\par{\noindent \bf Whole pages.}} %
The conclusions drawn from cleaned pages generally hold for whole pages as well, as shown in Table~\ref{tab:fullpageadditionalref_finetuned}. Since larger models have only medium QA quality on whole webpages,
we observed much higher quality gain with fine-tuning (up to 8\%), and multi-task fine-tuning improves further (6\%) over fine tuning with only QA tasks. On the other hand, incorporating triples extracted by 70+B models may degrade performance, likely due to their lower triple extraction accuracy on whole pages.

\smallskip
{\par{\noindent \bf Summary:}} We show that we can further improve QA accuracy by fine-tuning LLMs with triple extraction tasks, and the improvement is comparable to using triple augmentation. When combined with triple augmentation, we observe a quality boost (+11\% with ground truth triples on full webpages).

\section{Related Work} %
\label{sec:related}

\par{\noindent \bf Benchmarks.} Several publicly available benchmarks exist for semi-structured websites, including SWDE~\cite{hao2011from} and WEIR~\cite{bronzi2013extraction}, which focus on extracting predefined attributes of entities. In contrast, Expanded SWDE~\cite{lockard2019openceres} aims to recover all (subject, predicate, object) triples for \emph{detail webpages} that each contains information about a particular entity. Additionally, WebSRC~\cite{chen-etal-2021-websrc} is a dataset aimed at answering questions about semi-structured webpages. Our enriched WebSRC dataset distinguishes itself by covering both knowledge triple extraction and question answering tasks.

\smallskip

\par{\noindent \bf Approaches.} Early works have explored directly providing an answer to user questions by knowledge extraction from webpages~\cite{mausam2016open}, from distantly-supervised knowledge extraction~\cite{dong2014knowledge,lockard2018ceres} to GNN-based zero-shot knowledge extraction~\cite{lockard-etal-2020-zeroshotceres}. The success was capped by our capability of precisely extracting knowledge and our capability of leveraging the extracted knowledge which can contain errors. In contrast, we have observed more success in leveraging web tables~\cite{cafarella2018ten}, which takes the retrieval approach to return a webtable to answer the user question, bypassesing the hard retrieval and summarization steps. In recent years, there has been significant progress in knowledge extraction and question answering with the help of pre-trained language models~\cite{xie2021webke,zhao-etal-2022-tie,deng2022domlmlearninggeneralizablerepresentations,wang2022webformerwebpagetransformerstructure,zhang2024joint,hong2024combininglanguagegraphmodels} and LLMs~\cite{sarkhel2023self,tan2025htmlrag}. Following this trend and aiming to shed insights on this evolving landscape of web-based QA, we are the first to assess whether knowledge triple extraction can still provide value for QA over semi-structured content in the era of LLMs.

Another line of research has focused on leveraging accessible structured data sources, such as KGs, tables, or databases, for QA~\cite{jiang-etal-2023-structgpt}, as well as employing direct multimodal LLMs that process semi-structured data presented as images~\cite{faysse2025colpaliefficientdocumentretrieval,he-etal-2024-webvoyager}. In contrast, our work is motivated by the standard web-based RAG setting, where the semi-structured data arises from retrieval results without relying on additional dedicated, cleaned artifacts like databases, nor framing the problem as one of image or multimodal understanding.

\section{Conclusion} %
\label{sec:conclusion}

Our study demonstrates that while LLMs excel at QA over semi-structured web data, knowledge extraction---via triple augmentation and multi-task learning---still provides measurable benefits, especially for smaller models. However, the incremental value diminishes for state-of-the-art LLMs on cleaned webpages, highlighting the need for further research on real-world, noisier whole webpages. Improving extraction quality remains essential for broader applications beyond QA, such as knowledge integration and interpretability.

\bibliographystyle{ACM-Reference-Format}
\bibliography{latex/custom}

\appendix

\clearpage
\section{Appendix}
\label{sec:appendix}

\subsection{List of prompts}
\label{sec:prompt}

\begin{prompt}[ht!]
\centering
\small
\caption{Prompt employed in computing LLM-based QA metrics.}
\begin{tabular}{p{0.45\textwidth}}
\toprule
\textbf{<system>}\\
You need to check whether the prediction of a question-answering system to a question is correct. You should make the judgment based on the ground truth answer provided to you.
Your response should be "correct" if the prediction is correct or \"incorrect\" if the prediction is wrong. \\
\textbf{<user>}\\
Question: \{question\} \\
Ground truth: \{ground\_truth\} \\
Prediction: \{prediction\} \\
Correctness: \\
\bottomrule
\end{tabular}
\label{prompt:qametrics}

\end{prompt}

\begin{prompt}[ht!]
\centering
\small
\caption{Prompt employed in computing LLM-based triple extraction metrics.}
\begin{tabular}{p{0.45\textwidth}}
\toprule
\textbf{<system>}\\
You are given two (subject, predicate, object) triples. Your response should be "Yes" if the triples are semantically the same or "No" if they are semantically different. \\
\textbf{<user>}\\
\{triple\_1\} \\
\{triple\_2\} \\
\bottomrule
\end{tabular}
\label{prompt:temetrics}
\end{prompt}

\begin{prompt}[ht!]
\centering
\small
\caption{Triple extraction prompt.}
\begin{tabular}{p{0.45\textwidth}}
\toprule
\textbf{<system>}\\
You are given a doc in HTML and its title. Please return all (subject, predicate, object) triples that can be extracted from the doc, in the order they appear in the doc. Subject, predicate, and object should generally be gained from the text spans in the doc or the title. Please only include complete triples; if for any section the predicate or object is missing from the doc, you may skip it. Each line in your response should be a triple.\\
\textbf{<user>}\\
\#\#\# title\\
\{title\}\\
\#\#\# HTML\\
\{html\}\\
\bottomrule
\end{tabular}

\label{prompt:tezeroshot}

\end{prompt}

\begin{prompt}[ht!]
\centering
\small
\caption{Triple extraction prompt with demonstrations.}
\begin{tabular}{p{0.45\textwidth}}
\toprule
\textbf{<system>}\\
You are given a doc in HTML and its title. Please return all (subject, predicate, object) triples that can be extracted from the doc, in the order they appear in the doc. Subject, predicate, and object should generally be gained from the text spans in the doc or the title. Please only include complete triples; if for any section the predicate or object is missing from the doc, you may skip it. Each line in your response should be a triple.\\
\# Example 1\\
\#\#\# Input\\
title:\\
\{title\_1\}\\
HTML:\\
\{html\_1\}\\
\#\#\# Output\\
\{triples\_1\}\\
\# Example 2\\
\#\#\# Input\\
title:\\
\{title\_2\}\\
HTML:\\
\{html\_2\}\\
\#\#\# Output\\
\{triples\_2\}\\
\textbf{<user>}\\
\#\#\# title\\
\{title\}\\
\#\#\# HTML\\
\{html\}\\
\bottomrule
\end{tabular}

\label{prompt:tefewshot}

\end{prompt}

\begin{prompt}[ht!]
\centering
\small
\caption{Prompt employed by \textit{GenerateScriptFromLLM} in Algorithm~\ref{alg:scriptgeneration} \#\ref{alg:line:callprompt1}.}
\begin{tabular}{p{0.45\textwidth}}
\toprule
\textbf{<system>}\\
Your task is to write a program following the instruction. Your response should be a Python function(html: str) only without extra words. \\
\textbf{<user>}\\
Please write a Python function parse(html: str) to extract all (subject, predicate, object) triples from the html. \\
The return value should be a list of triples, in the order they appear in the html. \\
Subject, predicate, and object should generally be gained from the text spans in the doc. \\ 
The return value should only include complete triples; if for any section the predicate or object is missing from the doc, it may be skipped. \\
Below is an sample of Input and Output \\
\# Input (html) \\
<head><title>\{title\}</title></head> \\
\{html\} \\
\# Output (triples) \\
\{triples\} \\
\bottomrule
\end{tabular}

\label{prompt:scriptgen1}

\end{prompt}

\begin{prompt}[ht!]
\centering
\small
\caption{Prompt employed by \textit{GenerateScriptFromLLM} in Algorithm~\ref{alg:scriptgeneration} \#\ref{alg:line:callprompt2}.}
\begin{tabular}{p{0.45\textwidth}}
\toprule
\textbf{<system>}\\
Your task is to write a program following the instruction. Your response should be a Python function(html: str) only without extra words.\\
\textbf{<user>}\\
Please write a Python function parse(html: str) to extract all (subject, predicate, object) triples from the html.\\
The return value should be a list of triples, in the order they appear in the html.\\
Subject, predicate, and object should generally be gained from the text spans in the doc.\\ 
The return value should only include complete triples; if for any section the predicate or object is missing from the doc, it may be skipped. \\
Below are two samples of Input and Output\\
\# Sample 1\\
\#\# Input (html)\\
<head><title>\{title\_1\}</title></head>\\
\{html\_1\}\\
\#\# Output (triples)\\
\{triples\_1\}\\
\# Sample 2\\
\#\# Input (html)\\
<head><title>\{title\_2\}</title></head>\\
\{html\_2\}\\
\#\# Output (triples)\\
\{triples\_2\}\\
\bottomrule
\end{tabular}

\label{prompt:scriptgen2}

\end{prompt}

\begin{prompt}[ht!]
\centering
\small
\caption{Prompt employed by \textit{GenerateScriptFromLLM} in Algorithm~\ref{alg:scriptgeneration} \#\ref{alg:line:callprompt3}.}
\begin{tabular}{p{0.45\textwidth}}
\toprule
\textbf{<system>}\\
Your task is to fix/improve a program following the instruction if possible. Your response should be a Python function(html: str) only without extra words.\\
\textbf{<user>}\\
Please fix/improve a Python function parse(html: str) to extract all (subject, predicate, object) triples from the html.\\
The return value should be a list of triples, in the order they appear in the html.\\
Subject, predicate, and object should generally be gained from the text spans in the doc.\\ 
The return value should only include complete triples; if for any section the predicate or object is missing from the doc, it may be skipped. \\
Below is an sample of Input and Output\\
\# Input (html)\\
<head><title>\{title\}</title></head>\\
\{html\}\\
\# Output (triples)\\
\{triples\}\\
Here is the function and its execution result given the sample input:\\
\# Function\\
\{previous\_script\}\\
\# Execution result\\
\{execution\_result\}\\
\bottomrule
\end{tabular}

\label{prompt:scriptgen3}

\end{prompt}

\begin{prompt}[ht!]
\centering
\small
\caption{Prompt for QA with reference.}
\begin{tabular}{p{0.45\textwidth}}
\toprule
\textbf{<system>}\\
You are given a question and a reference that may or may not help answer the question. Please answer the question. Be concise. \\
\textbf{<user>}\\
\#\#\# Question\\
\{question\}\\
\#\#\# Reference\\
\{reference\}\\
\bottomrule
\end{tabular}

\label{prompt:qa1}

\end{prompt}

\begin{prompt}[ht!]
\centering
\small
\caption{Prompt for QA without reference.}
\begin{tabular}{p{0.45\textwidth}}
\toprule
\textbf{<system>}\\
Please answer the question. Be concise.\\
\textbf{<user>}\\
\#\#\# Question\\
\{question\}\\
\bottomrule
\end{tabular}

\label{prompt:vanillaqa}

\end{prompt}

\begin{prompt}[ht!]
\centering
\small
\caption{Prompt for QA with demonstrations.}
\begin{tabular}{p{0.45\textwidth}}
\toprule
\textbf{<system>}\\
You are given a question and a reference that may or may not help answer the question. Please answer the question. Be concise.\\
\# Example 1\\
\#\#\# Question\\
\{question\_1\}\\
\#\#\# Reference\\
\{reference\_1\}\\
\#\#\# Output\\
\{answer\_1\}\\
\# Example 2\\
\#\#\# Question\\
\{question\_2\}\\
\#\#\# Reference\\
\{reference\_2\}\\
\#\#\# Output\\
\{answer\_2\}\\
\textbf{<user>}\\
\#\#\# Question\\
\{question\}\\
\#\#\# Reference\\
\{reference\}\\
\bottomrule
\end{tabular}

\label{prompt:qa2}

\end{prompt}

\clearpage

\begin{table*}[ht!]
\small
\centering
\caption{Global FM and triple-level F-1\textsubscript{LM} across different layouts on cleaned pages. All numbers are in percentage (\%). }
\begin{tabular}{lllrrrrrr}
\toprule
& \bf Backbone & \bf Setting & \multicolumn{3}{c}{\bf Global FM} & \multicolumn{3}{c}{\bf Triple-level F-1\textsubscript{LM}} \\
& & & A-V & Hz & F-F & A-V & Hz & F-F  \\
\cmidrule(lr){1-6} \cmidrule(lr){7-9}

\multirow{10}{*}{in-domain} 
 & Llama 3.2-3B-Instruct & zero-shot & 54.5 & 51.8 & 41.4 & 42.5 & 39.6 & 27.6 \\ %
 & Llama 3.2-3B-Instruct & 2-shot & 54.0 & 44.4 & 45.8 & 62.9 & 66.1 & 49.7 \\ %
 & Llama 3.2-3B-Instruct & fine-tuned & 82.4 & 60.9 & 60.3 & 71.7 & 53.0 & 57.1 \\ %
\cmidrule(lr){2-6} \cmidrule(lr){7-9}

& Llama 3.1-70B-Instruct & zero-shot & 62.6 & 65.6 & 44.8 & 64.3 & 76.0 & 40.4 \\ %
& Llama 3.1-70B-Instruct & 2-shot & 95.3 & 84.9 & 88.6 & 92.4 & 85.3 & 91.4 \\ %
& Llama 3.1-70B-Instruct & fine-tuned & 89.3 & 81.8 & 61.9 & 83.2 & 80.1 & 56.5 \\ %
 & Claude 3.7 Sonnet & 2-shot & 96.0 & 92.8 & 87.6 & 93.4 & 91.6 & 88.1 \\ %
 & GPT-4o & 2-shot & 97.5 & 91.4 & 92.1 & 96.5 & 92.2 & 94.6 \\ %
\cmidrule(lr){2-6} \cmidrule(lr){7-9}

 & generated scripts & single call & 67.3 & 64.7 & 43.3 & 55.7 & 57.4 & 39.7 \\ %
 & generated scripts & multiple calls with feedback & 84.8 & 75.9 & 60.3 & 85.1 & 73.2 & 56.0  \\ %
\cmidrule(lr){1-6} \cmidrule(lr){7-9}

\multirow{10}{*}{out-of-domain} 
& Llama 3.2-3B-Instruct & zero-shot & 66.1 & 42.5 & 43.0 & 54.8 & 24.3 & 32.1 \\ %
& Llama 3.2-3B-Instruct & 3-shot &  62.3 & 40.0 & 35.8 & 53.1 & 13.4   & 21.3  \\ %
& Llama 3.2-3B-Instruct & fine-tuned & 84.4 & 56.3 & 53.0 & 74.0 & 35.0 & 54.2  \\ %
\cmidrule(lr){2-6} \cmidrule(lr){7-9}

& Llama 3.1-70B-Instruct & zero-shot & 66.7 & 60.2 & 45.0 & 71.1 & 68.9 & 51.1 \\ %
& Llama 3.1-70B-Instruct & 3-shot & 86.0 & 69.2 & 68.8 & 83.5 & 67.5 & 56.1  \\  %
& Llama 3.1-70B-Instruct & fine-tuned & 85.8 & 73.8 & 66.5 & 76.2 & 72.7 & 63.2  \\ %
& Claude 3.7 Sonnet & 3-shot & 87.2 & 79.2 & 65.0 & 87.0 & 80.6 & 53.9  \\ %
& GPT-4o & 3-shot & 87.6 & 75.4 & 61.8 & 89.8 & 76.6 & 58.0  \\ %
\cmidrule(lr){2-6} \cmidrule(lr){7-9}
 & generated scripts & single call & 78.4 & 54.5 & 43.2 & 64.5 & 46.4 & 30.1  \\
 & generated scripts & multiple calls with feedback & 82.7 & 57.3 & 55.2 & 81.0 & 56.2 & 43.5 \\
\bottomrule
\end{tabular}

\label{tab:tripleextractionlayout}
\end{table*}

\begin{table*}[t!]
\centering
\small
\caption{In-domain QA performance in Accuracy\textsubscript{LM} (\%) across different layouts on cleaned pages. Script-extracted triples enhance QA results with the most pronounced improvement observed in horizontal layout.}
\begin{tabular}{llrrrrrrrr}
\toprule
 \bf Setting & \bf Additional reference & \multicolumn{4}{c}{\bf L-3B} & \multicolumn{4}{c}{\bf Q-3B} \\
 & & \bf All & \bf A-V & \bf Hz & \bf F-F & \bf All & \bf A-V & \bf Hz & \bf F-F \\
\cmidrule(lr){1-6} \cmidrule(lr){7-10}
\multirow{3}{*}{zero-shot} & / & 77.1 & 85.9 & 72.4 & 80.0 & 81.5 & 88.1 & 79.9 & 80.4  \\ %
& Script-extracted triples & 80.6 & 87.6 & 79.2 & 78.9 & 87.5 & 89.2 & 87.8 & 85.9  \\ %
\cmidrule(lr){2-6} \cmidrule(lr){7-10}
 & Ground truth triples & 89.1 & 91.2 & 88.8 & 88.5 & 89.2 & 89.8 & 91.4 & 85.1  \\ %
\cmidrule(lr){1-6} \cmidrule(lr){7-10}
\multirow{3}{*}{2-shot} & / & 75.6 & 91.2 & 69.3 & 77.5 & 83.1 & 95.0 & 78.7 & 83.7  \\ %
&  Script-extracted triples  & 82.9 & 89.8 & 83.5 & 77.8 & 86.8 & 95.0 & 87.1 & 81.6 \\ %
\cmidrule(lr){2-6} \cmidrule(lr){7-10}
 & Ground truth triples & 85.6 & 88.7 & 85.4 & 84.0 & 89.4 & 94.8 & 90.3 & 84.7  \\ %
\bottomrule
\end{tabular}

\label{tab:qascriptextractedtripleslayout}
\end{table*}

\begin{table*}[t!]
\centering
\small
\caption{In-domain QA performance in Accuracy\textsubscript{A} (\%) on cleaned pages. }
\begin{tabular}{llrrrrr}
\toprule
 \bf Setting & \bf Additional reference & \bf L-3B & \bf L-70B & \bf Q-3B & \bf Q-72B \\
\midrule
\multirow{2}{*}{zero-shot} & / & 63.9 & 81.7 & 63.9 & 81.4  \\ %
& Script-extracted triples & 70.6 & 84.6 & 72.1 & 82.9   \\ %
\cmidrule(lr){1-6}
\multirow{2}{*}{2-shot} & / & 69.8 & 92.3 & 79.6 & 93.0 \\ %
&  Script-extracted triples & 75.3 & 92.7 & 81.5 & 94.2 \\ %
\bottomrule
\end{tabular}

\label{tab:indomainqaacca}
\end{table*}

\begin{table*}[t!]
\small
\centering
\caption{Out-of-domain QA performance in Accuracy\textsubscript{A} (\%) on cleaned pages.} 
\begin{tabular}{lrrrrr}
\toprule
\bf FT task & \bf L-3B & \bf L-70B & \bf Q-3B & \bf Q-72B \\
\cmidrule(lr){1-5}
/    & 65.2 & 87.8 & 67.7 & 88.4  \\ %
QA  & 79.9 & 95.9 & 80.1 & 93.7 \\ %
TE  & 68.5 & 88.1  & 63.0 & 89.5 \\ %
QA + TE  & 84.4 & 97.2 & 82.7 & 94.2  \\ %
\bottomrule
\end{tabular}

\label{tab:multitaskqaacca}
\end{table*}

\begin{table*}[t!]
\small
\centering
\caption{Out-of-domain QA performance in Accuracy\textsubscript{A} (\%) on cleaned pages. }
\begin{tabular}{llrrrr}
\toprule
\bf Fine-tuning task & \bf Additional reference & \bf L-3B & \bf L-70B & \bf Q-3B & \bf Q-72B \\
\cmidrule(lr){1-4} \cmidrule(lr){5-6}
\multirow{4}{*}{QA} & / & 79.9  & 95.9  &  80.1 & 93.7  \\  %
 & FT 3B extracted triples & 77.3  & / &  79.8  & /\\ %
 & FT 70+B extracted triples & 85.0 & 94.2 & 82.0 & 94.4 \\ %
 & Ground truth triples & 93.2 & 94.8  &  89.9 &  93.1 \\ %
\cmidrule(lr){1-4} \cmidrule(lr){5-6} 
\multirow{4}{*}{QA + Triple Extraction} & / & 84.4  & 97.2 & 82.7  & 94.2  \\ %
 & FT 3B extracted triples & 83.2  & / & 80.8  & /\\ %
& FT 70+B extracted triples &  87.5  &  95.4 & 83.7 & 94.9 \\ %
& Ground truth triples & 94.0  & 96.2  & 90.8  & 93.6  \\ %
\bottomrule
\end{tabular}

\label{tab:additionalref_finetunedacca}
\end{table*}

\begin{table*}[ht!]
\small
\centering
\caption{Triple extraction performance on cleaned pages. All numbers are in percentage (\%).}
\begin{tabular}{lllrrrrrrr}
\toprule
& \bf Backbone & \bf Setting & \multicolumn{7}{c}{\bf Triple-level} \\
& & & EM & P\textsubscript{EM} & R\textsubscript{EM} & F-1\textsubscript{EM} & P\textsubscript{FM} & R\textsubscript{FM} & F-1\textsubscript{FM} \\
\cmidrule(lr){1-4} \cmidrule(lr){5-7} \cmidrule(lr){8-10}

\parbox[t]{2mm}{\multirow{10}{*}{\rotatebox[origin=c]{90}{in-domain}}}  & Llama 3.2-3B-Instruct & zero-shot & 0.4 & 0.4 & 2.0 & 0.6 & 41.4 & 65.3 & 50.7 \\ %

& Llama 3.2-3B-Instruct & 2-shot & 42.4 & 42.4 & 61.2 & 50.1 & 60.2 & 87.8 & 71.4 \\ %
 & Llama 3.2-3B-Instruct & fine-tuned & 40.7 & 47.6 & 45.5 & 46.5 & 74.6 & 77.9 & 76.2 \\ %
\cmidrule(lr){2-4} \cmidrule(lr){5-7} \cmidrule(lr){8-10}
& Llama 3.1-70B-Instruct & zero-shot & 8.0 & 8.5 & 12.3 & 10.1 & 54.3 & 76.0 & 63.4 \\ %
& Llama 3.1-70B-Instruct & 2-shot & 75.3 & 77.7 & 81.7 & 79.6 & 90.3 & 92.8 & 91.5 \\ %
& Llama 3.1-70B-Instruct & fine-tuned & 50.6 &  52.7 & 54.3 & 53.5 & 78.6 & 82.7 & 80.6   \\ %
 & Claude 3.7 Sonnet & 2-shot & 79.8 & 80.5 & 83.4 & 81.9 & 92.9 & 94.9 & 93.9  \\ %
 & GPT-4o & 2-shot & 81.8 & 83.9 & 83.0 & 83.5 & 93.8 & 94.1 & 94.0  \\ %
 \cmidrule(lr){2-4} \cmidrule(lr){5-7} \cmidrule(lr){8-10}
 & generated scripts & single call & 30.9 & 58.8 & 32.0 & 41.5 & 57.5 & 52.7 & 55.0 \\ %
 & generated scripts & multiple calls with feedback & 56.1 & 73.6 & 56.9 & 64.2 & 81.1 & 75.6 & 78.2   \\ %

\cmidrule(lr){1-4} \cmidrule(lr){5-7} \cmidrule(lr){8-10}
\parbox[t]{2mm}{\multirow{8}{*}{\rotatebox[origin=c]{90}{out-of-domain}}} 
& Llama 3.2-3B-Instruct & zero-shot & 4.5 & 4.5 & 6.6 & 5.3 & 44.2 & 58.8 & 50.5\\ %
& Llama 3.2-3B-Instruct & 3-shot & 9.3 & 15.5 & 10.3 & 12.4 & 37.5 & 35.7 & 36.6 \\ %
& Llama 3.2-3B-Instruct & fine-tuned & 26.7 & 29.4 & 28.9 & 29.2 & 70.8 & 66.3 & 68.5   \\ %
\cmidrule(lr){2-4} \cmidrule(lr){5-7} \cmidrule(lr){8-10}
& Llama 3.1-70B-Instruct & zero-shot & 8.8 & 9.4 & 11.4 & 10.3 & 60.3 & 73.5 & 66.3 \\ %
& Llama 3.1-70B-Instruct & 3-shot & 42.1 & 45.3 & 46.2 & 45.7 & 76.5 & 76.2 & 76.4 \\  %
& Llama 3.1-70B-Instruct & fine-tuned & 41.1 & 45.3 & 42.9 & 44.1 & 79.1 & 77.0 & 78.0  \\ %
\cmidrule(lr){2-4} \cmidrule(lr){5-7} \cmidrule(lr){8-10}

& Claude 3.7 Sonnet & 3-shot & 48.1 & 49.6 & 51.2 & 50.4 & 80.8 & 83.2 & 82.0  \\ %
& GPT-4o & 3-shot & 40.1 & 41.7 & 42.5 & 42.1 & 78.1 & 78.5 & 78.3  \\ %

\bottomrule
\end{tabular}

\label{tab:tripleextractionallextended}
\end{table*}

\begin{table*}[t!]
\centering
\small
\caption{Zero-shot QA accuracy (\%) on cleaned pages.} %
\begin{tabular}{lrrrrr}
\toprule
\bf Reference & \multicolumn{2}{c}{\bf Accuracy\textsubscript{A}} & \multicolumn{2}{c}{\bf Accuracy\textsubscript{LM}} \\
 & L-3B & L-70B & L-3B & L-70B  \\
\midrule
\textit{(None)}  & 3.6 & 7.4  & 4.4  & 11.5  \\ %
HTML & 62.8 & 88.0 & 69.2 & 94.0 \\ %
Flattened & 65.2 & 87.8 & 72.6 & 94.7 \\ %
\bottomrule
\end{tabular}

\label{tab:qazeroshot}  %
\end{table*}

\clearpage
\subsection{Discussion of alternative metrics}
\label{sec:supplementalresults}

Compared to LLM-based metrics, rule-based metrics offer the advantages of being more deterministic and having lower computational costs. Therefore, we define and report additional rule-based metrics for both QA and triple extraction tasks.

\paragraph{QA accuracy.} We report \textbf{Appearance-based Accuracy} \textbf{(Accuracy\textsubscript{A})}, where the correctness is determined by whether the ground truth answer appears within the first 50 tokens of the model's response after normalization (\ie, lowercase, space normalization, remove leading and ending punctuations).

Table~\ref{tab:multitaskqaacca} and Table ~\ref{tab:additionalref_finetunedacca} report the QA performance in Accuracy\textsubscript{A}, which correspond to %
Table~\ref{tab:multitaskqa} and Table~\ref{tab:additionalref_finetuned}, respectively. Overall, we can observe that the performance measured in Accuracy\textsubscript{LM} is generally higher than Accuracy\textsubscript{A}, especially for the zero-shot setting. This is because LLMs, particularly when not fine-tuned, occasionally provide a correct response that does not strictly contain the ground truth answer as a substring. This suggests that the real performance can be underestimated by Accuracy\textsubscript{A}. Nevertheless, Accuracy\textsubscript{A} aligns with Accuracy\textsubscript{LM} in terms of the relative performance and conclusions.

\paragraph{Triple extraction quality.} Let $A$ be the set of triples in the prediction and $B$ be the set of triples in the ground truth, we define exact match-based metrics as follows, where we consider two triples are the same if subject, predicate, and object are the same after normalization (\ie, lower-casing, space normalization, and punctuation stripping). 

\begin{itemize}
\item \textbf{Exact Match (EM)}:= $|A\cap B|/\max(|A|,|B|)$ 
\item \textbf{EM-based Precision (P\textsubscript{EM})}:= $|A\cap B|/|A|$
\item \textbf{EM-based Recall (R\textsubscript{EM})}:= $|A\cap B|/|B|$
\item \textbf{EM-based F-1 Score (F-1\textsubscript{EM})}:= $2\cdot\text{P\textsubscript{EM}}\cdot\text{R\textsubscript{EM}}/(\text{P\textsubscript{EM}}+\text{R\textsubscript{EM}})$ 
\end{itemize}

Additionally, we define \textbf{Fuzzy Match-based Precision P\textsubscript{FM}}, \textbf{Recall (R\textsubscript{FM})}, and \textbf{F-1 Score (F-1\textsubscript{FM})}. These metrics are analogous to their LLM-based counterparts, with the exception that we utilize the mean character-level fuzzy match scores of the triple pairs from the maximum weight matching as the numerator in calculating precision and recall, rather than relying on the matching rate verified by the LLM.

Table~\ref{tab:tripleextractionallextended} shows the triple extraction performance in the exact match-based and fuzzy match-based metrics, which corresponds to Table~\ref{tab:tripleextractionall}. 

\subsection{Implementation details}
\label{sec:implementationdetail}

\paragraph{Training and inference.}  When the target task is triple extraction, we pass the webpage in HTML to LLMs as input. For question answering and multi-task training, we use the flattened page for page representation obtained by Beautiful Soup\footnote{\url{https://www.crummy.com/software/BeautifulSoup/}.}, as our preliminary study (Table~\ref{tab:qazeroshot}) showed that it yields slightly better performance compared to using HTML directly. Additionally, our preliminary study (Table~\ref{tab:qazeroshot}) validated that the answers in our experiments are mainly generated through RAG, as LLMs without references (Prompt~\ref{prompt:vanillaqa}) achieve an accuracy lower than 12\%.

We generally used Prompt~\ref{prompt:tezeroshot} and Prompt~\ref{prompt:tefewshot} for triple extraction, and Prompt~\ref{prompt:qa1}, Prompt~\ref{prompt:vanillaqa}, and Prompt~\ref{prompt:qa2} for question answering. These prompts were employed with nuanced adaptations as needed (\eg, add an additional demonstration according to the particular experimental setting). In particular, when using the flattened page as a reference, the reference was set to be the concatenation of the title (provided separately in the WebSRC dataset) and the flattened text (returned by Beautiful Soup after passing the HTML to it).

We leveraged Transformers~\cite{wolf-etal-2020-transformers}  and Unsloth~\cite{unsloth} for supervised fine-tuning on A100 (40G) or H100 (80GB) GPUs. In all fine-tuning experiments, we employed QLoRA~\cite{dettmers2023qloraefficientfinetuningquantized} with settings r=16 and alpha=32, and utilized prompt-masking. Fine-tuning was performed for 1 epoch, with a batch size of 2, and learning rates of 2e-4 for 3B models and 5e-5 for 70+B models.

\paragraph{Evaluation.} Our LLM-based metrics involve a two-stage process. First, we apply straightforward rules to assess clear-cut cases (\ie, the triples match exactly, the response matches perfectly the ground truth answer). For the remaining cases, we rely on LLMs, employing Prompt~\ref{prompt:qametrics} for QA metrics and Prompt~\ref{prompt:temetrics} for triple extraction metrics.

\subsection{Additional annotation details}
\label{sec:annotationdetails}

\begin{figure*}[!h]
\centering
  \begin{subfigure}[b]{\textwidth}
  \centering
  \includegraphics[width=0.7\linewidth]{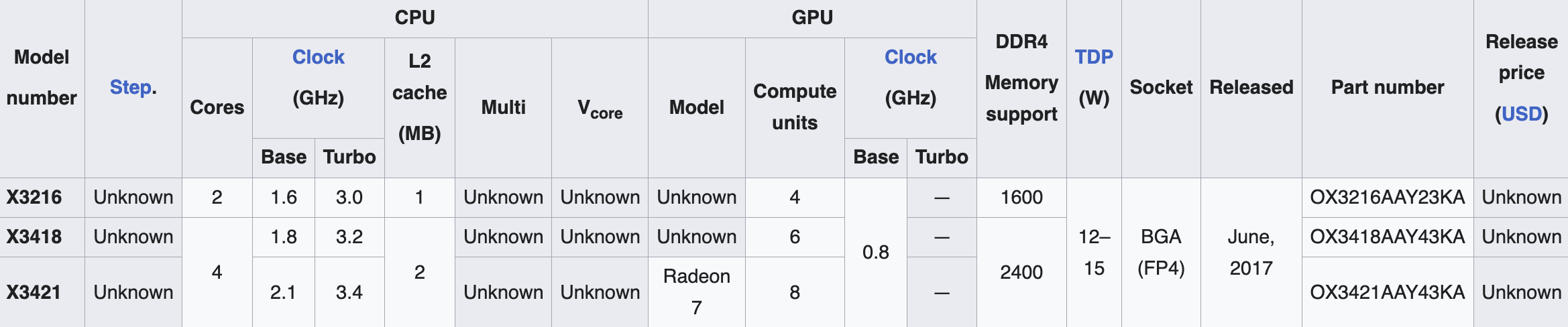} %
  \label{fig:horizontal}
  \end{subfigure}
  \begin{subtable}[b]{0.75\textwidth}
  \centering
  \small
    \begin{tabular}{lll}
    \\
    \toprule
    \bf Subject & \bf Predicate & \bf Object \\
    \midrule
    \multicolumn{3}{c}{\bf $\cdots$} \\
X3216 & \textit{<-- CPU Clock (GHz) -->} Base & 1.6 \\
\cmidrule(lr){1-3} 
    \multicolumn{3}{c}{\bf $\cdots$} \\
X3216 & \textit{<-- GPU Clock (GHz) -->} Base & 0.8 \\
    \multicolumn{3}{c}{\bf $\cdots$} \\
    \bottomrule
    \end{tabular}
  \end{subtable}
\caption{A triple annotation example with disambiguation information.}
\label{fig:annotationdisambiguation}
\end{figure*}

For the webpage shown in Figure~\ref{fig:annotationdisambiguation}, predicates such as ``CPU'' and ``GPU'' do not have corresponding objects. As a result, they do not form complete triplets and can be skipped. However, there are two instances of the predicate ``Base'' with different meanings. In such cases, we may add disambiguation information to the predicates to clarify their context.

\subsection{Details of QA pair construction for whole webpages}
\label{sec:fullpageqapairgen}
To prevent generating questions solely about content that the LLMs easily interpret, an initial set of candidate QA pairs was generated by prompting a 70B Llama model with the webpage content and \emph{corresponding ground truth triples}. We then refined the set with the LLMs by: (1) Removing questions that require heavy reasoning, as we focus on evaluating the ability to comprehend semi-structured webpages rather than heavily reasoning over them. (2) Removing questions that combined multiple distinct questions into one (\eg, ``What is an atom and how big is it?'') to avoid artificially increasing difficulty by combining multiple questions. (3) Filtering out questions that were correctly answered by all LLMs from the set \{Llama 3.1-70B-Instruct, Llama 3.1-8B-Instruct, GPT-4o, GPT-4o mini, Claude 3.7 Sonnet, Gemma 3-27b-it~\cite{gemmateam2025gemma3technicalreport}, Phi-3.5-MoE-instruct~\cite{abdin2024phi3technicalreporthighly}\} to create questions that effectively differentiate model performance. Finally, we performed human audits to eliminate QA pairs where questions were not grounded in the webpage content or where answers were incorrect.

\begin{table*}[t!]
\small
\centering
\caption{Ablation tests for Algorithm~\ref{alg:scriptgeneration} on cleaned pages.}
\begin{tabular}{lrrrrr}
\toprule
 & \multicolumn{1}{c}{\bf Global} & \multicolumn{4}{c}{\bf Triple-level} \\
 & FM & EM & P\textsubscript{EM} & R\textsubscript{EM} & F-1\textsubscript{EM} \\
\cmidrule(lr){1-2} \cmidrule(lr){3-6}
multiple calls with feedback, 3 iterations (\ie, Algorithm~\ref{alg:scriptgeneration}) & 74.5 & 56.1 & 73.6 & 56.9 & 64.2 \\ %
multiple calls with feedback, 2 iterations (\ie, change 3 to 2 in \#\ref{alg:line:foriterbegin})  & 74.2 & 55.1 & 72.6 & 55.9 & 63.2 \\ %
multiple calls with feedback, 1 iteration (\ie, change 3 to 1 in \#\ref{alg:line:foriterbegin}) & 74.8 & 51.3 & 67.1 & 52.3 & 58.8   \\ %
multiple calls with no feedback (\ie, remove \#\ref{alg:line:foriterbegin}--\#\ref{alg:line:foriterend}) & 70.2 & 42.1 &  62.1 & 42.6 & 50.5 \\ %
single call (\ie, remove \#\ref{alg:line:forsamplebegin}--\#\ref{alg:line:forsampleend}) & 59.0 & 30.9 & 58.8 & 32.0 & 41.5 \\ %

\bottomrule
\end{tabular}

\label{tab:teablation}
\end{table*}

\begin{algorithm*}[ht!]
  \caption{GenerateScript(A, B)}
  \label{alg:scriptgeneration}
  \begin{algorithmic}[1]
    \Require Two samples $A$ and $B$, each containing HTML and the ground truth triples
    \State Initialize collection of candidate scripts: $\mathit{scriptCands} \gets \{\}$
    \State Generate script using both samples and add to candidate scripts: $\mathit{scriptCands} \gets \mathit{scriptCands} \cup \{\text{GenerateScriptFromLLM}(samples = \{A, B\})\}$  \label{alg:line:callprompt2}
    \For{each sample $s \in \{A, B\}$} \label{alg:line:forsamplebegin} 
      \State Generate sample-specific base script: $\mathit{script}_0 \gets \text{GenerateScriptFromLLM}(samples = \{s\})$ \label{alg:line:callprompt1}
      \State Add base script to candidate scripts: $\mathit{scriptCands} \gets \mathit{scriptCands} \cup \{\mathit{script}_0\}$
      \For{$i \gets 1$ to $3$} \label{alg:line:foriterbegin}
        \State Execute previous script on sample's HTML: $\mathit{execResultPrev} \gets \text{ExecuteScript}(script = \mathit{script}_{i-1}, html = s.html)$
        \State Generate improved script using feedback: $\mathit{script}_i \gets \text{GenerateScriptFromLLM}(samples = \{s\}, previousScript = \mathit{script}_{i-1}, execResult = \mathit{execResultPrev})$ \label{alg:line:callprompt3}
        \State Add improved script to candidate scripts: $\mathit{scriptCands} \gets \mathit{scriptCands} \cup \{\mathit{script}_i\}$
      \EndFor \label{alg:line:foriterend}
    \EndFor \label{alg:line:forsampleend}
    \State \Return $\text{SelectBestScript}(\mathit{scriptCands}, evaluationSamples = \{A, B\})$
  \end{algorithmic}
\end{algorithm*}

\subsection{Details of triple extraction through automatically generated scripts}
\label{sec:ablationscript}
Algorithm~\ref{alg:scriptgeneration} describes the steps of generating a script for in-domain triple extraction through multiple calls with feedback, where \emph{GenerateScriptFromLLM} in \#\ref{alg:line:callprompt2} \#\ref{alg:line:callprompt1} and \#\ref{alg:line:callprompt3} employs Prompt~\ref{prompt:scriptgen2}, Prompt~\ref{prompt:scriptgen1}, and Prompt~\ref{prompt:scriptgen3}, respectively. For fast, low-cost iteration, we employed exact match defined in Appendix~\ref{sec:supplementalresults} as the metric for script selection in \emph{SelectBestScript}. We report ablation tests in Table~\ref{tab:teablation}. The out-of-domain triple extraction follows the same algorithm, except that it uses triples extracted by the 3-shot extraction model (based on Llama 3.1-70B-Instruct) instead of ground truth triples.

\end{document}